\documentclass[twoside]{article}
\usepackage[accepted]{aistats2018}
\usepackage{graphicx}
\usepackage{subfigure}
\usepackage{bm,bbm}
\usepackage{mathrsfs}
\usepackage{algorithm}
\usepackage{algorithmic}

\renewcommand{\algorithmicensure}{ \textbf{Initialize:}}

\usepackage{amsfonts}
\usepackage{amssymb,amsthm}
\usepackage[cmex10]{amsmath}
\usepackage{epstopdf}
\usepackage{color}
\usepackage{setspace}
\newtheorem{theorem}{Theorem}
\newtheorem{corollary}[theorem]{Corollary}
\newtheorem{assumption}[theorem]{Assumption}
\newtheorem{proposition}[theorem]{Proposition}
\newtheorem{lemma}[theorem]{Lemma}
\allowdisplaybreaks[4]

\begin{document}
\twocolumn[
\aistatstitle{Regional Multi-Armed Bandits}
\aistatsauthor{ Zhiyang Wang, Ruida Zhou,  Cong Shen }
\aistatsaddress{School of Information Science and Technology \\University of Science and Technology of China\\ \textsf{\{wzy43, zrd127\}@mail.ustc.edu.cn, congshen@ustc.edu.cn}}  
]

\begin{abstract}

We consider a variant of the classic multi-armed bandit problem where the expected reward of each arm is a function of an unknown parameter. The arms are divided into different groups, each of which has a common parameter. Therefore, when the player selects an arm at each time slot, information of other arms in the same group is also revealed. This regional bandit model naturally bridges the \emph{non-informative bandit} setting where the player can only learn the chosen arm, and the \emph{global bandit} model where sampling one arms reveals information of all arms. We propose an efficient algorithm, \textsf{UCB-g}, that solves the regional bandit problem by combining the Upper Confidence Bound (UCB) and greedy principles. Both parameter-dependent and parameter-free regret upper bounds are derived. We also establish a matching lower bound, which proves the order-optimality of \textsf{UCB-g}. Moreover, we propose  \textsf{SW-UCB-g}, which is an extension of \textsf{UCB-g} for a non-stationary environment where the parameters slowly vary over time. 
\end{abstract}

\section{Introduction}
\label{sec:intro}

Multi-armed bandit (MAB) is a useful tool for online learning. The player can choose and play one arm from a set of arms at each time slot. An arm, if played, will offer a reward that is drawn from its distribution which is unknown to the player. The player's goal is to design an arm selection policy that maximizes the total reward it obtains over finite or infinite time horizon.  MAB is a basic example of sequential decision with an exploration and exploitation tradeoff \cite{Bubeck:12}.

The classic MAB setting focuses on {independent} arms, where the random rewards associated with different arms are independent. Thus, playing an arm only reveals information of this particular arm. This \emph{non-informative bandit} setting is a matching model for many applications, and has received a lot of attention \cite{LaiRobbins,Auer:02,Bubeck:12}. However, in  other applications, the statistical rewards of different arms may be {correlated}, and thus playing one arm also provides information on some other arms. A \emph{global bandit} model has been studied in \cite{Atan:15}, where the rewards of all arms are (possibly nonlinear and monotonic) functions of a common unknown parameter, and all the arms are correlated through this global parameter. As a result, sampling one arm reveals information of all arms. It has been shown that such dependency among arms can significantly accelerate the policy convergence, especially when the number of arms is large \cite{Atan:15}.

The \emph{non-informative} and \emph{global} bandits lie on the two opposite ends of the informativeness spectrum. Their fundamental regret behaviors and optimal arm selection polices are also drastically different. In this work, we aim at bridging these two extremes by studying a new class of MAB model, which we call \emph{regional bandits}.  In this new model, the expected reward of each arm remains a function of a single parameter. However, only arms that belong to the same group share a common parameter, and the parameters across different groups are not related. The agent knows the functions but not the parameters. By adjusting the group sizes and the number of groups, we can smoothly shift between the non-informative and global bandits extremes. The policy design and regret analysis of the regional bandits model thus can provide a more complete characterization of the whole informativeness spectrum than the two extreme points.



Regional bandit is a useful model for some real world problems, such as dynamic pricing with demand learning and market selection, and drug selection/dosage optimization. In dynamic pricing with an objective of maximizing revenue over several markets, the agent sequentially selects a price $p\in\mathcal{P}$ in market $m$ at time $t$ and observes sales $S_{p,t}(\theta_m)=(1-\theta_mp)^2+\epsilon_{t}$ which is modeled in \cite{Huang:13} as a function of market size $\theta_m$, and $\epsilon_{t}$ is a random variable with zero mean. The revenue is then given by $R_{p,t}=p(1-\theta_mp)^2+p\epsilon_{t}$. In this example, the market sizes $\{\theta_m\}$ are the regional parameters which stay constant in the same market and need to be estimated by setting different prices and observing the corresponding sales.

In drug dosage optimization, the dosage(C)/effect(E) relationship is characterized by $\frac{E}{E_{\text{max}}}=\frac{C}{K_D+C}-E_0$ in \cite{Holford:81}, where $K_D$ is a parameter for medicine category $D$ and $C$ is the dosage concentration. By using different medicine with different dosage levels, the effect of dosage can be learned. $K_D$ can be seen as the regional parameters that need to be estimated in  group $D$.

We make the following contributions in this work.
\begin{enumerate}
\item We propose a parametric regional bandit model for group-informative MABs and a \textsf{UCB-g} policy, and derive both parameter-dependent and parameter-free upper bounds for the cumulative regret. By varying the group size and the number of groups, the proposed model, the  \textsf{UCB-g} policy, and the corresponding regret analysis provide a complete characterization over the entire informativeness spectrum, and incorporate the two extreme points as special cases.
\item We prove a parameter-dependent lower bound for the regional bandit model, which matches the regret upper bound of \textsf{UCB-g} and proves its order optimality.
\item We further study a non-stationary regional bandit model, where the parameters of each group may change over time. We propose a sliding-window-based  \textsf{UCB-g}  policy, named  \textsf{SW-UCB-g}, and prove a time-averaged regret bound which depends on the drifting speed of the parameter.
\item We adopt a practical dynamic pricing application and perform numerical experiments to verify the performance of the proposed algorithms.
\end{enumerate}

\section{Related Literature}
\label{sec:review}

There is a large amount of literature on MAB problems. We focus on the literature that is related to the informativeness among arms.

\subsection{Non-informative and Global Bandits}
For the standard bandits with independent arms (i.e., \emph{non-informative} bandits), there are a great amount of literature including the ground-breaking work of Lai and Robbins \cite{LaiRobbins} for finite-armed stochastic bandits.  The celebrated UCB algorithm was proposed in \cite{Auer:02}, which provides a $O(K\log T)$ regret bound where $K$ is the number of arms and $T$ is the finite time budget.  For adversarial bandits, \cite{Auer:12} gave the EXP4 algorithm with a regret bounded by $O(\sqrt{TN\log K})$, where $N$ is the number of experts. The other extreme is the global bandit setting, where arms are related through a common parameter. In  \cite{Mersereau:09}, the authors considered a linear model and a greedy policy was proposed with a bounded regret. In  \cite{Atan:15} the model was extended to a generic function (possibly nonlinear) of a global parameter and the authors proposed a \emph{Weighted Arm Greedy Policy} (WAGP) algorithm, which  can also achieve a parameter-dependent bounded regret.

\subsection{Group-informative Bandits}
There are some existing works which has considered a similar \emph{group-informative} bandit setting as our regional bandits, but the model and algorithms are very different. In  \cite{Mannor:11}, based on a known graphic structure, additional side observations are captured when pulling an arm. This is done using unbiased estimates for rewards of the neighborhood arms of the selected arm. An exponentially-weighted algorithm with linear programming was proposed, whose regret is bounded by $O(\sqrt{c\log N T})$. The performance depends on the characteristics of the underlying graph, and some computational constraints exist. In  \cite{Cesa-Bianchi:12}, a combinatorial bandit structure was proposed. In this model, the player receives the sum reward of a subset of arms after pulling an arm, which can be seen as a specific case for the general linear bandit setting \cite{Auer:022}. A strategy was proposed whose regret bound is also sublinear in time.

Both these two works are constructed as an adversarial online learning problem. Different from these group-informative bandit models, our model focuses on the stochastic setting. Moreover, we adopt a parametric method and allow for general reward functions, in order to capture both individual and group reward behaviors. We also get rid of the need for extra side observations about other arms when one arm is played, which is impractical for some applications.


\section{Problem Formulation}
\label{sec:set}

There are $M$ groups of arms, with arm set $\mathcal{K}_m:=\{1,..,K_m\}$ for group $m\in\mathcal{M}$. The expected rewards of arms in group $m$ depend on a single parameter $\theta_m\in \Theta_m$, while arms in different groups are not related, i.e. the elements in vector $\bm{\theta}=[\theta_1,\theta_2,..,\theta_M]$ are unstructured. For ease of exposition, we normalize $\Theta_m$ as a subset of the unit interval $[0,1]$. The expected reward of an arm $k\in\mathcal{K}_m$ is a known invertible function of $\theta$, denoted as $\mu_{m,k}(\theta)$. Therefore an arm can be completely determined by the two indices $[m,k]$. Each time this arm is pulled, a reward $X_{m,k}(t)$ is revealed which is a random variable drawn from an unknown distribution $\nu_{m,k}(\theta_m)$, with $\mathbb{E}_{\nu_{m,k}(\theta_m)}[X_{m,k}(t)]=\mu_{m,k}(\theta_m)$. Above all, the parameters set $\bm{\theta}$ together with the reward functions $\{\mu_{m,k}(\theta)\}$ define the regional bandit machine.

Without further assumptions on the functions, the problem can be arbitrarily difficult. Therefore, some regularities need to be imposed as follows.

\begin{assumption}
\label{assu:1}
(i) For each $m \in \mathcal{M}$, $k \in \mathcal{K}_m$ and $\theta,\theta' \in \Theta_m,$ there exists
$D_{1,m,k} > 0$ and $1 < \gamma_{1,m,k}$, such that:
\begin{equation*}
|\mu_{m,k}(\theta)-\mu_{m,k}(\theta')| \geq D_{1,m,k}|\theta-\theta'|^{\gamma_{1,m,k}}.
\end{equation*}
(ii)For each $m \in \mathcal{M}$, $k \in \mathcal{K}_m$ and $\theta,\theta' \in \Theta_m,$ there exists
$D_{2,m,k} > 0$ and $0< \gamma_{2,m,k} \leq 1$, such that:
\begin{equation*}
|\mu_{m,k}(\theta)-\mu_{m,k}(\theta')| \leq D_{2,m,k}|\theta-\theta'|^{\gamma_{2,m,k}}.
\end{equation*}
\end{assumption}
The first assumption ensures the monotonicity of the function, while the second  is known as the \emph{H\"{o}lder continuity}. Naturally, these assumptions also guarantee the same properties for the inverse reward functions, as stated in Proposition \ref{pro:2}.

\begin{proposition}
\label{pro:2}
For each $m \in \mathcal{M}$, $k \in \mathcal{K}_m$ and $y, y' \in [0,1],$
\begin{equation*}
|\mu_{m,k}^{-1}(y)-\mu_{m,k}^{-1}(y')| \leq \bar{D}_{1,m,k}|y-y'|^{\bar{\gamma}_{1,m,k}}
\end{equation*}
under Assumption \ref{assu:1}, where $\bar{\gamma}_{1,m,k} = \frac{1}{\gamma_{1,m,k}}$, $\bar{D}_{1,m,k} = {(\frac{1}{D_{1,m,k}})}^{\frac{1}{\gamma_{1,m,k}}}$.
\end{proposition}

Proposition \ref{pro:2} and the invertibility indicate that the rewards we receive from a particular arm can be used to estimate the parameter $\theta_m$ of the group, therefore improving the estimate of expected rewards of other arms in the group. Notice that the H\"{o}lder continuous reward functions are possibly \emph{nonlinear}, which leads to biases in the estimations\footnote{This is a critical difference to the linear bandit model.} and must be handled in the estimation. We also define \textcolor{black}{$D_{1,m} = \min_{k\in \mathcal{K}_m}{D_{1,m,k}}$, $\gamma_{1,m} = \max_{k\in \mathcal{K}_m}{\gamma_{1,m,k}}$, $\bar{D}_{1,m}=\max_{k\in \mathcal{K}_m}{\bar{D}_{1,m,k}}$, $\bar{\gamma}_{1,m}=1/\gamma_{1,m}=\min_{k\in \mathcal{K}_m}{\bar{\gamma}_{1,m,k}}$, $D_{2,m} = \max_{k\in \mathcal{K}_m}{D_{2,m,k}}$ and $\gamma_{2,m} = \min_{k\in \mathcal{K}_m}{\gamma_{2,m,k}}$} for all $m\in\mathcal{M}$ and $k\in\mathcal{K}_m$.


In the regional bandit model, the player chooses one arm at each time slot based on previous observations and receives a random reward, drawn independently from the reward distribution of the chosen arm. The objective is to maximize the cumulative reward up to a time budget $T$. When complete knowledge of $\bm\theta$ is known by an omniscient play, the optimal arm, denoted by $[m^*,k^*]=\arg\max_{m\in\mathcal{M},k\in \mathcal{K}_m}\mu_{m,k}(\theta_m)$, would always be chosen. We denote this as the optimal policy and use it to benchmark the player's policy that selects arm $[m(t),k(t)]$ at time $t$, whose performance is measured by its regret:
\begin{equation*}
R(\bm{\theta},T)=T\mu^*(\bm{\theta})- \sum_{t=1}^T \mathbb{E}[\mu_{m(t),k(t)}(\theta_{m(t)}) ],
\end{equation*}
where $\mu^*(\bm{\theta})=\mu_{m*,k*}(\theta_{m^*})$.


\section{ Algorithm and Regret Analysis}
\label{sec:policy}

\subsection{The \textsf{UCB-g} Policy}
\label{sec:ucbg}

The proposed \textsf{UCB-g} policy combines two key ideas that are often adopted in bandit algorithms:  \emph{Upper Confidence Bound (UCB)}, and \emph{greediness}. More specifically, the \textsf{UCB-g} policy handles three phases separately in the regional bandit problem -- group selection, arm selection, and parameter update. The detailed algorithm is given in Algorithm \ref{alg:UGP}.

For the group selection phase, since groups are independent of each other with respect to the single parameter $\theta_m$, the $(\alpha,\psi)$-UCB method can be adopted \cite{Bubeck:12}. We establish the upper envelope function and suboptimality gap in the following proposition.

\begin{proposition}
\label{pro:3}
For each $m \in \mathcal{M}$, $k \in \mathcal{K}_m$ and $\theta,\theta' \in \Theta_m$, we denote $\mu_m(\theta)$ as the upper envelope function of the arms in group $m$. Therefore, $\mu_{m}(\theta_m)=\max\limits_{k\in\mathcal{K}_m}\mu_{m,k}(\theta_m)$ and there must exist $k\in \mathcal{K}_{m}$ that satisfies:
\begin{equation*}
|\mu_{m}(\theta)-\mu_{m}(\theta')| \leq |\mu_{m,k}(\theta)-\mu_{m,k}(\theta')|.
\end{equation*}
$\Delta_{m}= \mu_{m^*}(\theta_{m^*})-\mu_{m}(\theta_{m})$ is defined as the suboptimal gap of group $m$ compared to the group that contains the optimal arm.
\end{proposition}

Following the UCB principle, at each time step, the index of the chosen group is computed as the estimated reward plus a padding function, accounting for the uncertainty of the estimation. The padding function is defined as:
\begin{equation}
\textcolor{black}{\psi_m^{-1}(x)=D_{2,m}{\bar D_{1,m}}^{\gamma_{2,m}}(x)^{\xi_m},}
\label{eqn:padding}
\end{equation}
where \textcolor{black}{$\xi_m=\frac{\bar\gamma_{1,m}\gamma_{2,m}}{2}$} and $N_m(t)$ denotes the number of times arms in group $m$ are chosen up to time $t$. Note that the choice of padding function \eqref{eqn:padding} is non-trivial compared to the standard $(\alpha,\psi)$-UCB  \cite{Bubeck:12}.  The policy selects group $m(t)$ as follows:
\begin{equation*}
m(t) = \arg\max_{m \in \mathcal{M}} \mu_{m}(\hat\theta_m(t)) + \psi_m^{-1}\left(\frac{\alpha_m\log(t)}{N_m(t-1)}\right),
\end{equation*}
where $\hat{\theta}_m(t)$ is the estimated parameter of group $m$, $\alpha_m$ is a constant larger than $K_m$ and ties are broken arbitrarily. We can see that the form of the padding function has similar flavor to UCB but with a different exponent related to the characteristics of the functions. \textcolor{black}{We note that the chosen exponent guarantees convergence of the algorithm as will be proved in Theorem \ref{the:UG}. In standard UCB \cite{Bubeck:12}, the exponent of the padding function is generally set to 0.5; in our setting, however, $\bar{\gamma}_{1,m}$ and $\gamma_{2,m}$ are smaller than 1, thus our algorithm leads to a smaller exploration item because of the parameterized reward functions.}

\begin{algorithm}[h]
\caption{The \textsf{UCB-g} Policy for Regional Bandits}
\label{alg:UGP}
\begin{algorithmic}[1]
\begin{spacing}{1.3}
\REQUIRE $\mu_{m,k}(\theta)$ for each $m \in \mathcal{M}$, $k \in \mathcal{K}_m$\\
\algorithmicensure  {}  $ t=1, N_{m,k}(0)=0$ for each $k \in \mathcal{K}$\\
\WHILE{$t \leq T$}
\IF{ $t \leq M$}
\STATE{Select group $m(t)=t$ and randomly select arm $k(t)$ from set $\mathcal{K}_{m(t)}$}
\ELSE
\STATE{Select group $m(t) = \arg\max\limits_{m \in \mathcal{M}} \mu_{m}(\hat\theta_m(t)) + \psi_m^{-1}(\frac{\alpha_m\log(t)}{N_m(t-1)})$,
  and select arm $k(t) = \arg\max\limits_{k \in \mathcal{K}_{m(t)}}\mu_{m(t),k}(\hat\theta_{m}(t))$}
\ENDIF
\STATE Observe reward $X_{m(t),k(t)}(t)$
\STATE Set $\hat X_{m,k}(t)=\hat X_{m,k}(t-1)$, $N_{m,k}(t) = N_{m,k}(t-1)$ for all $m \neq m(t)$ and $k \neq k(t)$
\STATE $\hat X_{m(t),k(t)}(t)=$\\ $\frac{N_{m(t),k(t)}(t-1) \hat X_{m(t),k(t)}(t-1) + X_{m(t),k(t)}(t)}{N_{m(t),k(t)}(t-1) + 1}$, $N_{m(t),k(t)}(t) = N_{m(t),k(t)}(t-1) + 1$
\STATE $\hat k_m = \arg\max_{k}N_{k,m}(t)$ for all $m \in \mathcal{M}$
\STATE $\hat \theta_{m}(t)=\mu_{m,\hat k_m}^{-1}(\hat X_{m,\hat k_m}(t))$
\STATE $t=t+1$
\ENDWHILE
\end{spacing}
\end{algorithmic}
\end{algorithm}

After selecting the group, the next arm selection phase follows a greedy principle, which selects an arm with the highest estimated average reward in group $m(t)$, without adjusting for uncertainty:
\begin{equation*}
k(t) \in \arg\max\limits_{k \in \mathcal{K}_{m(t)}}\mu_{m(t),k}(\hat\theta_{m}(t)).
\end{equation*}

Finally, the player pulls  arm  $k(t) $ and receives a random reward $X_{m(t),k(t)}(t)$. The parameter update phase first updates the expected reward estimate of the selected arm. Then it uses the estimated reward of arm $\hat{k}_m$ to update the parameter estimate $\hat{\theta}_m(t)$.

\subsection{Regret Analysis}
\label{sec:perform}

We need to define some quantities to help with the analysis. In group $m$, each arm $k\in\mathcal{K}_m$ can be optimal for some $\theta\in\Theta_m$. We define the set of these $\theta$'s as the \emph{optimal region} for arm $[m,k]$, denoted as $\Theta^*_{m,k}$. Furthermore, we have $\Theta^*_{m,k}\neq \emptyset$; otherwise arm $[m,k]$ will never be selected in our greedy policy. $\theta_m$ denotes the true parameter of group $m$ and we define $\underline\Theta_m:=\bigcup\limits_{\theta_m\notin\Theta^*_{m,k^*}}\Theta^*_{m,k}$ as the suboptimal region for arm $[m,k]$.

\begin{figure}[h]
\centerline{\includegraphics[width=0.50\textwidth]{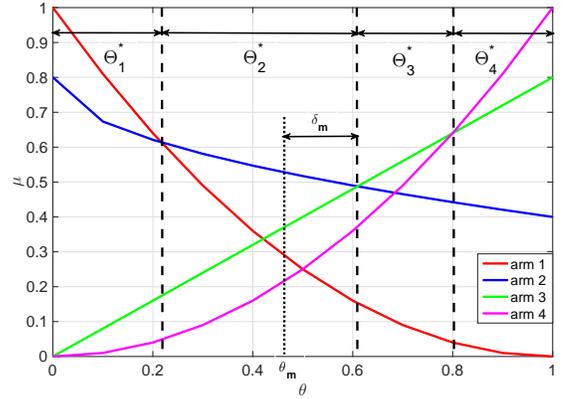}}
\caption{An illustration of suboptimal regions with 4 arms in a group, whose reward functions are $\mu_{m,1}(\theta)=(\theta-1)^2$, $\mu_{m,2}(\theta)=0.8-0.4\sqrt{\theta}$, $\mu_{m,3}(\theta)=0.8\theta$, $\mu_{m,4}(\theta)=\theta^2$.}
\label{fig:armmodel}
\end{figure}

To correctly select the best arm, the estimated $\hat{\theta}_m(t)$ should not fall in $\underline\Theta_m$. Therefore, we define the \emph{biased distance} $\delta_m=\min\{ |\theta_m-\theta|\}, \theta \in \underline\Theta_m$, which is the smallest distance between $\theta_m$ and the suboptimal region. A pictorial illustration is given in Fig.~\ref{fig:armmodel}. When the distance between the estimated $\hat{\theta}_m(t)$ and $\theta_m$ is within $\delta_m$, the policy would select the best arm and therefore optimal performance can be guaranteed. \textcolor{black}{We also denote $\delta=\delta_{m^*}$ as the biased distance in the optimal group.}

With these preliminaries, we first derive a finite-time parameter-dependent regret upper bound with $\delta$ and $\Delta_m$. The main result is given in Theorem \ref{the:UG}, which shows a sublinear regret of the proposed policy for the regional bandit model. Because of the independence of rewards across groups, as will be seen in Section~\ref{sec:low}, this logarithmic behavior of the regret is unavoidable. However, since the UCB principle is only applied to group selection, the \textsf{UCB-g} policy performs especially well when the number of groups is small compared to the total number of arms.

\begin{theorem}
\label{the:UG}
Under Assumption \ref{assu:1}, with $\alpha_m>K_m$ the regret of \textsf{UCB-g} policy is bounded as:
\begin{align}
\label{eqn:UG}
R(\bm{\theta},T) & \leq  \sum_{m \neq m^*}(\frac{\alpha_m\log(T)}{\psi_m(\Delta_{m}/2)}+\frac{2}{\alpha-2} ) \nonumber \\
& + \frac{2\left(1-\exp(-\frac{2T}{K_{m^*}}(\frac{\delta}{\bar D_{1,m^*}})^{2\gamma_{1,m^*}})\right)}{\exp(\frac{2}{K_{m^*}}(\frac{\delta}{\bar D_{1,m^*}})^{2\gamma_{1,m^*}}) -1} \\
 &=O(\log(T)), \nonumber
\end{align}
with $\alpha=\max_m 2\alpha_m/K_m$.
\end{theorem}

Theorem~\ref{the:UG} is important as it characterizes the regret bound for any group size and number of groups, hence covers the entire informativeness spectrum. It is natural to look at the two extreme points of the spectrum, whose regret upper bounds are known. More specifically,
the following corollary shows that Theorem~\ref{the:UG} covers them as special cases.

\begin{corollary}
\begin{enumerate}
\item With $M=1$, bound~\eqref{eqn:UG} becomes
\begin{align*}
R(\theta,T)&=R_C(T)\leq \frac{2\left(1-\exp(-\frac{2T}{K}(\frac{\delta}{\bar D_{1}})^{2\gamma_{1}})\right)}{\exp(\frac{2}{K}(\frac{\delta}{\bar D_{1}})^{2\gamma_{1}}) -1}
\\&\leq \frac{2}{\exp(\frac{2}{K}(\frac{\delta}{\bar D_{1}})^{2\gamma_{1}}) -1}
\end{align*}
which coincides with the result in  \cite{Atan:15}. 
\item When $K_1=\cdots=K_M=1$, bound~\eqref{eqn:UG} becomes
\begin{equation*}
R(\theta,T)=R_B(T)\leq \sum_{m \neq m^*}\left(\frac{\alpha_m\log(T)}{\psi_m(\Delta_{m}/2)}+\frac{2}{\alpha-2}\right)
\end{equation*}
which is consistent with the result in \cite{Bubeck:12}.
\end{enumerate}
\end{corollary}

Our next main result is the \emph{worst-case regret} of the \textsf{UCB-g} algorithm, given in the following theorem.
\begin{theorem}
\label{the:indereg}
Under Assumption \ref{assu:1}, the worst-case regret of \textsf{UCB-g} policy is:
\begin{equation*}
 \sup\limits_{\bm{\theta}_*\in\Theta}R(\bm{\theta}_*,T) \leq C_1(M\log T)^{\xi}T^{1-\xi}+C_2K_{m^*}^{\xi_{m^*}}T^{1-\xi_{m^*}},
\end{equation*}
where $\xi=\max_{m\in\mathcal{M}} \xi_m$ representing the worst case.
\end{theorem}

The worst-case performance is sublinear in time horizon as well as the number of groups and arms in the optimal group. Also, it is a parameter-free bound.


\section{Lower Bounds}
\label{sec:low}

In this section we show that the performance of the \textsf{UCB-g} policy is essentially unimprovable in logarithmic order due to the independence between groups. The loss is caused by selecting a suboptimal arm and therefore we will bound the number of times suboptimal arms are chosen. As we have noted in the previous section, a suboptimal arm may be either in the optimal group or in the suboptimal group. Therefore we will analyze the number of plays for them separately.

First, we present the lower bound on the worst-case regret in the following theorem.
\begin{theorem}
\label{the:low-worst}
\begin{equation*}
\sup\limits_{\bm\theta_*\in\Theta}R(\bm\theta_*,T)=\Omega \left(\min\{\sqrt{MT}, T^{1-\xi}\} \right).
\end{equation*}
\end{theorem}

Next, we focus on the parameter-dependent lower bound. The main result is given in the following theorem.
\begin{theorem}
\label{thm:deplow}
Without loss of generality, we assume that $\mu_1(\theta_1)>\mu_2(\theta_2)..\geq\mu_M(\theta_M)$. We also assume that the reward distributions are identifiable\footnote{The probability measures satisfy $p_0\neq p_1, 0<K(p_0,p_1)<+\infty$.}, and for all $\beta\in(0,1]$ there exists a strategy that satisfies $R(T)=o(T^\beta)$. The number of selections of suboptimal arms can be lower bounded as:
\begin{align}
\label{eqn:low2}
\nonumber 
& \mathbb{E}(N_{\mathcal{A}}(T))\geq \sum_{a\in\mathcal{A}_1}\frac{1}{4\textsf{KL}(\mu_{1,*}(\theta_1),\mu_{1,a}(\theta_1))}
\\&+\sum_{m=2}^M\left(\frac{1}{\textsf{KL}_{\inf}(\mu_m(\theta_m);\mu_1(\theta_1))}+o(1)\right)\log(T),
\end{align}
where $\mathcal{A}_1=\mathcal{K}_1/[1,*]$ stands for the arms in group 1 that are not optimal. $\textsf{KL}(p_0,p_1)$ is the Kullback--Leibler divergence between two probability measures $p_0$ and $p_1$, where $p_1$ is absolutely continuous with respect to $p_0$, and $\textsf{KL}_{\inf}(p_0, p_1) \doteq \inf\{\textsf{KL}(p_0, q):\mathbb{E}_{X\sim q}>p_1\}$.
\end{theorem}
For the first part of \eqref{eqn:low2}, we will show that the number of plays in a particular group can be lower bounded by a parameter-dependent constant. For simplicity, we only prove the case of two arms, but the result can be easily generalized to $K$ arms. The ideas are similar to \cite{Bubeck:13}. We first rephrase the arm selection problem as a \emph{hypothesis testing}, and invoke the following well-known lower bound for the minimax risk of hypothesis testing \cite{Tsyabkov:09}.
\begin{lemma}
\label{lem:hypo}
Let $\Psi$ be a maximum likelihood test:
\begin{equation*}
\Psi=
\begin{cases}
0, p_0\geq p_1,\\
1, p_0<p_1,
\end{cases}
\end{equation*}
where $p_0$ and $p_1$ are the densities of $P_0$ and $P_1$ with respect to $X$. Then we have
$$\mathbb{P}_{X\sim p_0}(\Psi(X)=1)+\mathbb{P}_{X\sim p_1}(\Psi(X)=0)\geq \frac{1}{2}e^{(-K(p_0,p_1))}.$$
\end{lemma}
Next we consider a simple two-armed case with reward functions $\mu_1(\theta)$ and $\mu_2(\theta)$, where the two arms have the same performance when $\theta=\theta_*$. We  assume, without loss of generality, that $\mu_1(\theta)$ is monotonically decreasing while $\mu_2(\theta)$ is increasing. Optimal regions are $\Theta^*_{1}=[0,\theta_*]$ and $\Theta^*_{2}=[\theta_*,1]$, respectively. We assume the rewards follow normal distributions for simplicity, and consider the case where arm 1 is optimal, i.e., $\theta\in\Theta^*_1$.
\begin{lemma}
\label{lem:subarm}
\begin{equation*}
\lim\limits_{T\rightarrow\infty}\mathbb{E}(N_2(T))\geq \frac{1}{4K(\mu_1(\theta),\mu_2(\theta))},
\end{equation*}
\end{lemma}

We now analyze selecting suboptimal arms from the suboptimal groups. The technical challenge is that our methodology must be different from the existing approach (such as in \cite{Kaufmann:16}), where the optimal arm is \emph{sequentially} switched. An important observation for the regional bandit model is that the performance change of one arm leads to changes of other arms in the same group. To circumvent this issue, we consider the group behavior and use the number of times a suboptimal group is selected to substitute for the number of times arms in this particular suboptimal group are selected. We have the following result.

\begin{lemma}
\label{lem:subgroup}
Without loss of generality, we assume that $\mu_1(\theta_1)>\mu_2(\theta_2)..\geq\mu_M(\theta_M)$. We also assume that the reward distributions are identifiable, and for all $\beta\in(0,1]$ the cumulative reward satisfies $R(T)=o(T^\beta)$. The number of selections of suboptimal group can be lower bounded as:
\begin{equation*}
\lim\limits_{T\rightarrow\infty}\frac{\mathbb{E}(N_m(T))}{\log(T)}\geq\frac{1}{K_{inf}(\mu_m(\theta_m);\mu_1(\theta_1))}.
\end{equation*}

\end{lemma}
The proof is similar to \cite{Kaufmann:16} and will be omitted. Putting Lemma~\ref{lem:subarm} and \ref{lem:subgroup} together immediately leads to Theorem~\ref{thm:deplow}.

Finally, a straightforward examination reveals that the developed lower bounds degenerate to known results in \cite{Atan:15} (for global bandits) and \cite{LaiRobbins} (for standard MAB), thus covering the two extreme cases.

\section{Non-stationary Regional Bandits}
\label{sec:nonsta}

We extend the regional bandit model to an environment where the parameter of each group $\theta_m$ may slowly change over time. In particular, the parameter for group $m$ is denoted as $\theta_m^t$ which varies with $t$. The random reward of arm $[m,k]$, $X_{m,k}(t)$, has a time-varying distribution with mean $\mu_{m,k}(\theta_m^t)$. We assume that the parameters vary smoothly. Specifically, they are assumed to be Lipschitz continuous.

\begin{assumption}
\label{assu:speed}
$\theta_m^t$ is Lipschitz continuous, i.e., for any $t$ and $t'$,
\begin{equation*}
|\theta_m^t-\theta_m^{t'}|\leq \left|\frac{t-t'}{\tau}\right|,
\end{equation*}
holds for all $m\in \mathcal{M}$, where $\tau>0$ controls the speed of drifting for the parameter.
\end{assumption}

As $\mu_m(\theta_m^t)$ for different groups $m$ may vary with time, it is possible that the rewards of two  groups may become very close to each other. As a result, estimate for the optimal group may be poor, which leads to a large regret due to selecting the suboptimal group.  Similar to  \cite[Assumption 1]{Combes:14}, we make an extra assumption to suppress such  circumstances in order to develop a performance-guaranteed algorithm.  We first define:
\begin{equation*}
G(\Delta,T)=\sum_{t=1}^T\sum_{m,m'\in\mathcal{M}}\mathbbm{1}_{|\mu_m(\theta_m^t)-\mu_{m'}(\theta_{m'}^t)|<\Delta},
\end{equation*}
as the confusing period. Then we have Assumption~\ref{assu:confuse}.
\begin{assumption}
\label{assu:confuse}
There exists a function $f$ and $\Delta_0$ such that for all $0\leq \Delta<\Delta_0$,
\begin{equation*}
\limsup_{T\rightarrow \infty} \frac{G(\Delta,T)}{T} \leq f(M)\Delta.
\end{equation*}
\end{assumption}

\subsection{\textsf{SW-UCB-g}}

A common approach to handle the non-stationarity in a bandit problem is to apply a sliding window (SW) on the observations \cite{Garivier:08}, which will keep the data ``fresh'' and thus eliminate the impact of obsolete observations. We follow the same idea and present the modified \textsf{UCB-g} strategy for the non-stationary setting in Alg.~\ref{alg:nonsta}. The basic operation follows  the stationary setting in Section \ref{sec:set}, with the main difference being that when estimating the parameter of each group,  only the latest $\tau_w$ observations are used. We also adopt a modified padding function $c_m(t,\tau_w)$, defined as:
\begin{equation*}
c_m(t,\tau_w)=D_{2,m}\bar{D}_{1,m}^{\gamma_{2,m}}\left(\frac{\alpha_m\log(t\wedge\tau_w)}{N_m(t,\tau_w)}\right)^{\xi_m},
\end{equation*}
where $t\bigwedge\tau_w$ represents the minimum of $t$ and $tau_w$, $N_m(t,\tau_w)$ denotes the number of times group $m$ is chosen in the past $\tau_w$ time slots before $t$, and $N_{m,k}(t,\tau_w)$ denotes the corresponding number of times arm $k$ in group $m$ is selected.


\begin{algorithm}[h]
\caption{The \textsf{SW-UCB-g} Policy for Regional Bandits with Non-stationary Parameters}
\label{alg:nonsta}
\begin{algorithmic}[1]
\begin{spacing}{1.3}
\REQUIRE $\mu_{m,k}(\theta)$ for each $k \in \mathcal{K}_m$, $m \in \mathcal{M}$\\
\algorithmicensure ${} t=1, N_{m,k}(0,\tau_h)=0, N_m(0,\tau_h)=0$ 
\WHILE{$t \leq T$}
\IF{ $t<M$}
\STATE{Select group $m(t)=t$ and randomly select arm $k(t)$ from set $\mathcal{K}_{m(t)}$}
\ELSE
\STATE{Select group $m(t) = \arg\max\limits_{m \in \mathcal{M}} \mu_{m}(\hat\theta_m(t))+c_m(t,\tau_w)$, and select arm $k(t) = \arg\max\limits_{k \in \mathcal{K}_{m(t)}}\mu_{m(t),k}(\hat\theta_{m}(t))$}
\ENDIF
\STATE Observe reward $X_{m(t),k(t)}(t)$
\STATE $\hat{X}_{m,k}(t,\tau_w)=\frac{\sum_{s=t-\tau_w+1}^tX_{m,k}(s)\mathbbm{1}_{\{m(s)=m,k(s)=k\}}}{N_{m,k}(t,\tau_w)}$, $N_{m,k}(t,\tau_w)=\sum_{s=t-\tau_w+1}^t\mathbbm{1}_{\{m(s)=m,k(s)=k\}}$

\STATE $\hat k_m = \arg\max\limits_{k\in\mathcal{K}_m}N_{m,k}(t,\tau_w)$, $\hat \theta_{m}(t)=\mu_{m,\hat k_m}^{-1}(\hat{X}_{m,k}(t,\tau_w))$ for $m \in \mathcal{M}$
\ENDWHILE
\end{spacing}
\end{algorithmic}
\end{algorithm}

\subsection{Regret Analysis}
Different from the stationary environments stated before, the change of parameters may lead to switches of the best arm. The regret here quantifies how well the policy tracks the best arm over time. We denote $[m^*(t),k^*(t)]$ as the optimal arm index at time $t$.  The cumulative regret up to time $T$   can be written as:
\begin{equation}
R(T)=\sum\limits_{t=1}\limits^{T}(\mu_{m^*(t),k^*(t)}(\theta_{m^*(t)}^t)-\mathbb{E}[\mu_{m(t),k(t)}(\theta_{m(t)}^t)]).
\end{equation}

\begin{theorem}
\label{the:nonsta}
Under Assumptions \ref{assu:1} and \ref{assu:speed}, with the window length set as $\tau_w=\max_{m\in\mathcal{M}}\tau^{\frac{2\gamma_{2,m}}{2\gamma_{2,m}+1}}$, the regret per unit time is:
\begin{equation}
\label{eqn:nonsta}
\lim_{T\rightarrow\infty}\frac{R(T)}{T}=O(\tau^{-\frac{\bar{\gamma}_1\gamma_2^2}{2\gamma_2+1}}+\tau^{-\frac{2\gamma_2}{2\gamma_2+1}} \log(\tau)),
\end{equation}
where $\gamma_2=\min\gamma_{2,m}, \bar{\gamma}_1=\max\bar\gamma_{1,m}$.
\end{theorem}

We see from Eqn.~\eqref{eqn:nonsta} in Theorem~\ref{the:nonsta} that the regret per unit time is a monotonically decreasing function of the speed $\tau$. It vanishes when $\tau\rightarrow\infty$, which is as expected since this corresponds to the case of stationary reward distributions.

\section{Numerical Experiments}
\label{sec:sim}

We carry out numerical simulations to compare \textsf{UCB-g} to UCB \cite{Auer:02} in a stationary setting, and \textsf{SW-UCB-g} to SW-UCB \cite{Garivier:08} in a non-stationary setting,  respectively. In addition to a basic experiment setting which uses the illustrative example of Fig.~\ref{fig:armmodel}, we also reported experiment results for a dynamic pricing application.

\subsection{Basic Experiment}

\begin{figure}
\centerline{\includegraphics[width=0.48\textwidth]{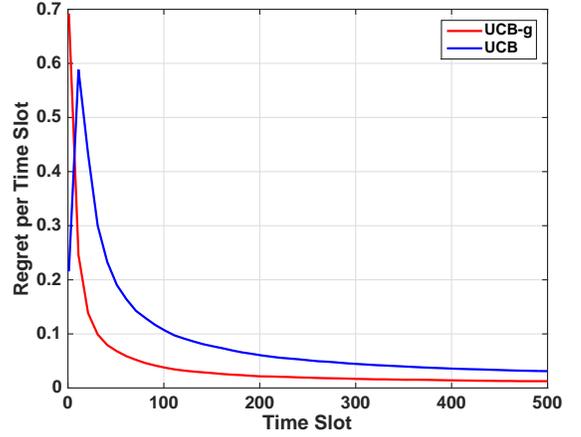}}
\caption{Regret per unit time in a stationary environment of the basic experiment setting.}
\label{fig:sta}
\end{figure}

In the first experiment, we consider $M=4$ groups and each group has $4$ arms. The reward functions remain the same as those used in Fig.~\ref{fig:armmodel}. The group parameters are set as $[\theta_1,\theta_2,\theta_3,\theta_4]=[0.1,0.4,0.7,1]$. We also have $\gamma_1=2,\gamma_2=0.5,D_1=0.1,D_2=2$. The comparison of per-time-slot regret of \textsf{UCB-g} and UCB is reported in Fig.~\ref{fig:sta}, which indicates that although both algorithms converge to the optimum asymptotically,  \textsf{UCB-g} outperforms UCB with lower regret. This is due to the exploitation of intra-group informativeness. 

\begin{figure}
\centerline{\includegraphics[width=0.43\textwidth]{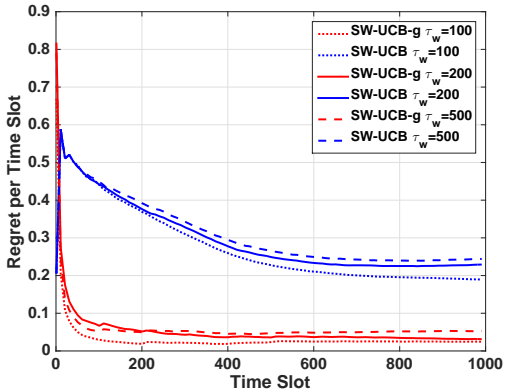}}
\caption{Regret per unit time in a non-stationary environment of the basic experiment setting.}
\label{fig:nonsta}
\end{figure}

For the non-stationary environment, we set the drifting speed $\tau=1000$, and the window size is set as $\tau_w=100, 200, 500$, respectively. The performances, measured by regret per unit time, are reported in Fig.~\ref{fig:nonsta}. We can see that \textsf{SW-UCB-g} has a much faster convergence than SW-UCB. Furthermore, we note that the regret performance is not monotonic with respect to the sliding window size $\tau_w$, e.g., $\tau_w=200$ is better than 500 but worse than 100 for large time budget $T$.


\begin{figure}
\centerline{\includegraphics[width=0.43\textwidth]{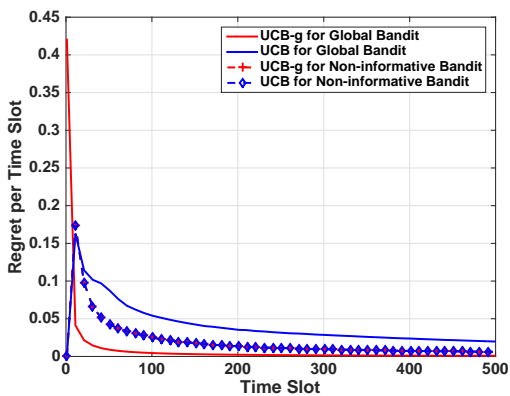}}
\caption{Regret per unit time for non-informative ($M=4$) and global bandits ($M=1$).}
\label{fig:onegrouponearm}
\end{figure}

As we have shown in the theoretical analysis, the \textsf{UCB-g} algorithm can recover the two extreme cases, non-informative MAB and global bandits, as special cases. We now verify this conclusion via simulations. If we change the group size to $M=1$ with 4 arms, we should recover the global bandit setting; if we change the group size to $M=4$ with 1 arm in each group, we should recover the standard non-informative bandit setting. The results are reported in Fig.~\ref{fig:onegrouponearm}. First, we can observe that \textsf{UCB-g} outperforms UCB when $M=1$. This is due to the exploitation of the common parameter by \textsf{UCB-g}. Next, we see that when $M=4$ with 1 arm in each group, \textsf{UCB-g} and UCB have identical performance, which is as expected.

%

\subsection{Example of Dynamic Pricing}

\begin{figure}[h]
\centerline{\includegraphics[width=0.43\textwidth]{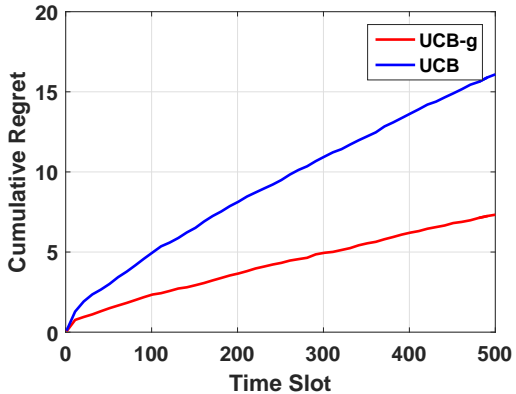}}
\caption{Cumulative regret for the dynamic pricing problem in a stationary environment.}
\label{fig:price}
\end{figure}

\begin{figure}[h]
\centerline{\includegraphics[width=0.43\textwidth]{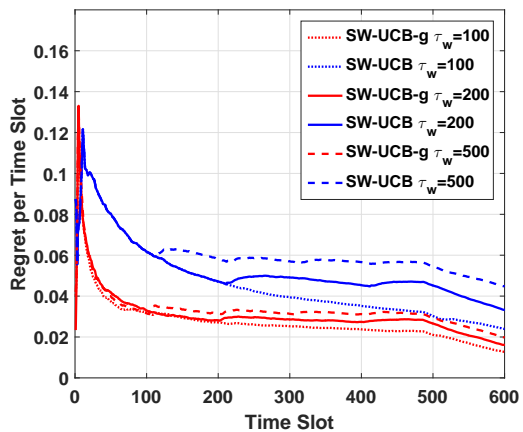}}
\caption{Regret per unit time for the dynamic pricing problem in a non-stationary environment.}
\label{fig:pricenon}
\end{figure}

For the dynamic pricing problem with demand learning and market selection, the expected revenue at time $t$ in market $m$ under price $p$ has the form $\mu_{m,p}(\theta_m)=\mathbb{E}[S_{p,t}(\theta_m)]=p(1-\theta_mp)^2.$ When selecting price $p$ in market $m$, the reward is generated from a standard Gaussian distribution. We set $\mathcal{K}_1=\{0.35,0.5\}$, $\mathcal{K}_2=\{0.35, 0.5, 0.7\}$, $\mathcal{K}_3=\{0.5, 0.7\}$, $\mathcal{K}_4=\{0.35, 0.5, 0.7,0.95\}$, $[\theta_1,\theta_2,\theta_3,\theta_4]=[0.35,0.5, 0.7,0.9]$, and then compare the proposed policy with UCB. The numerical result is presented in Fig \ref{fig:price}. Under a non-stationary environment, the change speed of the two market sizes is set to be $\tau=1000$ and the regret per unit time is reported in Fig.~\ref{fig:pricenon}. The same observations as in the basic experiment setting can be had from these results.

\section{Conclusion}
\label{sec:conc}
In this paper, we have addressed the stochastic bandit problem with a regional correlation model, which is a natural bridge between the non-informative bandit and the global bandit. We have proved an asymptotic lower bound for the regional model, and developed  the \textsf{UCB-g} algorithm that can achieve order-optimal regret by exploiting the intra-region correlation and inter-region independence.  We also extended the algorithm to handle non-stationary parameters, and proposed the \textsf{SW-UCB-g} algorithm that applies a sliding  window to the observations used in parameter estimation. We proved a bounded per-time-slot regret for \textsf{SW-UCB-g} under some mild conditions.  Simulation results have been presented to corroborate the analysis.

\clearpage

\subsubsection*{Acknowledgements}
This work has been supported by Natural Science Foundation of China (NSFC) under Grant 61572455, and the 100 Talent Program of Chinese Academy of Sciences.

\bibliographystyle{plain}
\bibliography{global}

\end{document}